\documentclass[runningheads]{llncs}

\usepackage{eccv}

\usepackage{graphicx}
\usepackage{booktabs}
\usepackage{multirow}
\usepackage[accsupp]{axessibility}  
\usepackage{bbding}
\usepackage{enumitem}
\usepackage[width=122mm,left=12mm,paperwidth=146mm,height=193mm,top=12mm,paperheight=217mm]{geometry}

\usepackage{hyperref}

\usepackage{orcidlink}

\def\aka{\emph{a.k.a.}}

\begin{document}

\setlength{\textfloatsep}{4pt}
\setlength{\floatsep}{4pt}
\setlength{\intextsep}{0pt}
\setlength{\abovecaptionskip}{4pt}
\setlength{\belowcaptionskip}{4pt}

\title{Learning Semantic Latent Directions for Accurate and Controllable Human Motion Prediction} 

\titlerunning{Semantic Latent Directions}

\author{
Guowei Xu\inst{1}$^{*}$ \and
Jiale Tao\inst{2}$^{*}$ \and
Wen Li\inst{1}$^{\dagger}$ \and
Lixin Duan
\inst{1}
}

\authorrunning{Xu et al.}

\institute{Shenzhen Institute for Advanced Study, \\
 University of Electronic Science and Technology of China
\and  
School of Computer Science and Engineering, \\ 
University of Electronic Science and Technology of China\\
\email{\{xuguowei368, jialetao.std, liwenbnu, lxduan\}@gmail.com}}

\maketitle

\renewcommand{\thefootnote}{\fnsymbol{footnote}}
\footnotetext[0]{$^{*}$ Guowei Xu and Jiale Tao contributed equally.\quad$\dagger$ Corresponding author.}

\begin{abstract}
  In the realm of stochastic human motion prediction (SHMP), researchers have often turned to generative models like GANS, VAEs and diffusion models. However, most previous approaches have struggled to accurately predict motions that are both realistic and coherent with past motion due to a lack of guidance on the latent distribution. In this paper, we introduce Semantic Latent Directions (SLD) as a solution to this challenge, aiming to constrain the latent space to learn meaningful motion semantics and enhance the accuracy of SHMP. SLD defines a series of orthogonal latent directions and represents the hypothesis of future motion as a linear combination of these directions. By creating such an information bottleneck, SLD excels in capturing meaningful motion semantics, thereby improving the precision of motion predictions. Moreover, SLD offers controllable prediction capabilities by adjusting the coefficients of the latent directions during the inference phase. Expanding on SLD, we introduce a set of motion queries to enhance the diversity of predictions. By aligning these motion queries with the SLD space, SLD is further promoted to more accurate and coherent motion predictions. Through extensive experiments conducted on widely used benchmarks, we showcase the superiority of our method in accurately predicting motions while maintaining a balance of realism and diversity. Our code and pretrained models are available at \href{https://github.com/GuoweiXu368/SLD-HMP/}{https://github.com/GuoweiXu368/SLD-HMP}.
  
  \keywords{Stochastic Human Motion Prediction \and Generative Models \and Semantic Latent Directions}
\end{abstract}

\section{Introduction}
\label{sec:intro}

Human motion prediction (HMP) aims to predict possible future motions from the observed pose sequence, with a wide range of applications on autonomous driving\cite{paden2016survey}, human-computer interaction\cite{koppula2015anticipating,interaction1,interaction2,interaction3,interaction4}, healthcare\cite{healthcare}, character animation\cite{character_animation}, and motion tracking\cite{luber2010people}. Despite recent progress on this topic\cite{Belfusion,Dlow,DivSamp,STARS,HumanMac,GSPS,MOJO,MotionDiff}, the nature of the underlying multimodal distribution of human motion, makes it still challenging to predict accurate and diverse motions.
\begin{figure}[tb]
  \centering
  \includegraphics[width=\textwidth]{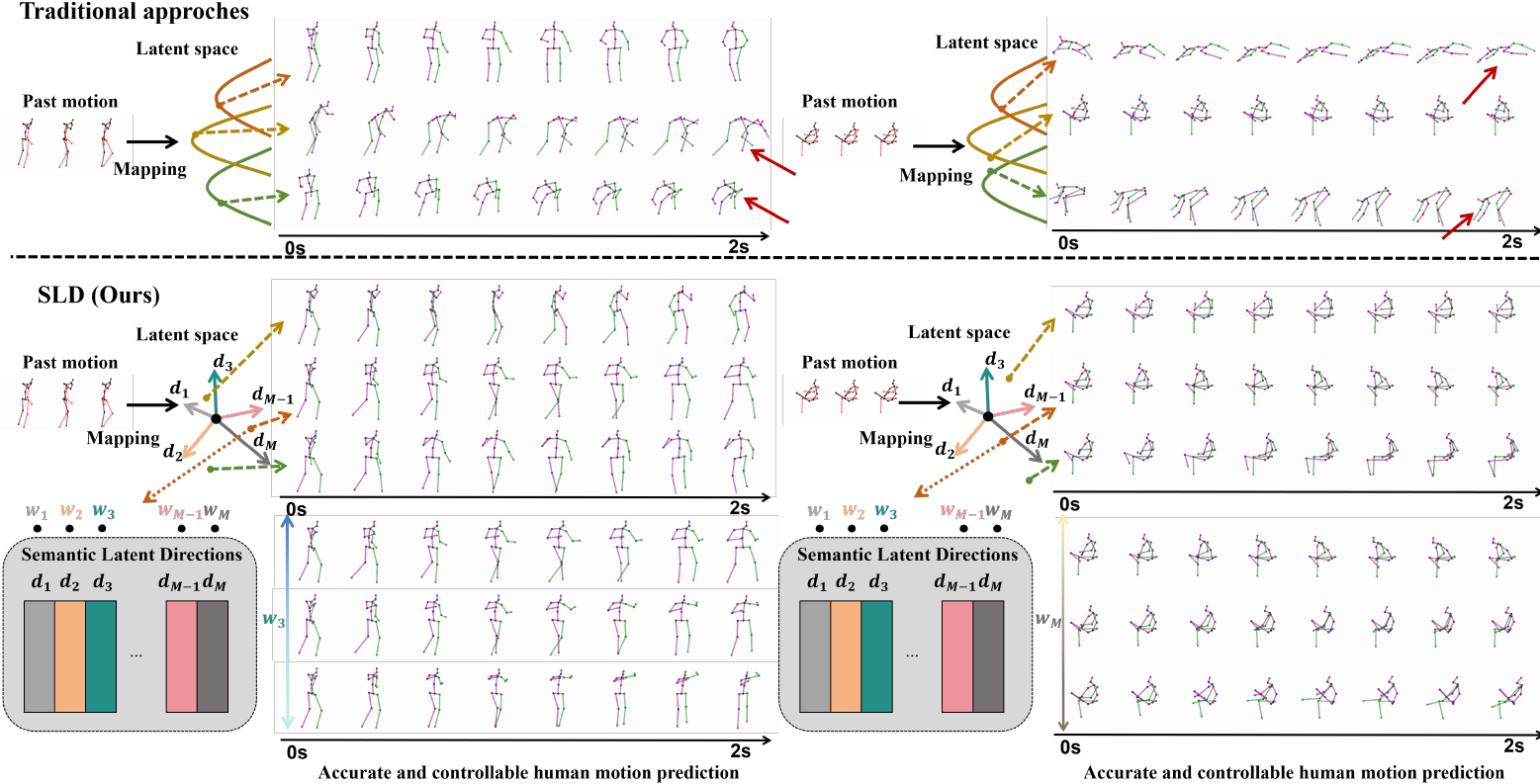}
  \caption{Traditional methods for stochastic human motion prediction typically involve learning a generative latent distribution without appropriate constraints and guidance. This often results in challenges in acquiring meaningful human motion representations, leading to inaccurate predictions characterized by abnormal poses and incoherent sequences compared to past motion patterns. In contrast, our proposed Semantic Latent Directions (SLD) framework leverages semantic latent directions to steer motion prediction, enabling the generation of future motions with high precision, realism, and coherence with past motion sequences. Moreover, SLD facilitates semantically controllable human motion prediction by adjusting the weights of the semantic latent directions, as illustrated in the bottom part.}
  \label{fig:intro}
\end{figure}

A commonly utilized approach\cite{Dlow,STARS,BOM,DivSamp,HumanMac,MotionDiff} in this field involves the development of robust generative models such as variational autoencoders, generative adversarial networks, and diffusion models. Taking variational autoencoders (VAEs) as an example, the encoder typically maps human motion to a latent distribution, often constrained by KL divergence to match a prior distribution like the Gaussian distribution. In predicting future motion, the VAE's decoder processes inputs from the embedding of past motion and a sampled vector from the prior distribution. While this approach establishes a solid generative framework for the Human Motion Prediction (HMP) task, the lack of appropriate constraints and guidance on the latent distribution hinders the models from learning meaningful human motion representations effectively. Consequently, these methods often struggle to predict accurate human motions, as they tend to focus on the major modes of the latent distribution with limited precision and coherence to past motion patterns, a phenomenon known as mode collapse. Some techniques, such as Dlow~\cite{Dlow} and STARS~\cite{STARS}, have introduced post-hoc or deterministic strategies for diverse motion sampling based on the learned latent distribution. However, even with these strategies, the models still face challenges in accurately and realistically predicting human motions due to the limited expressiveness of the learned latent distribution (Fig.~\ref{fig:intro}, upper).

In this paper, we aim to address the limitations mentioned above by seeking a suitable approach for latent motion representation that can not only encompass a wide range of hypotheses for prediction but also accurately capture the distribution of future motion. Our approach involves constructing a set of orthogonal bases in the latent space and representing hypothesis motion as a linear combination of these bases. To forecast future motion, we first predict the coefficients of the orthogonal bases and then decode the latent representation into a pose sequence. The constructed latent space offers several advantages. Firstly, by imposing orthogonal constraints on the bases, we can effectively model the diverse motion distribution of the dataset by aligning different coefficients with the bases. Secondly, unlike previous methods that often generate unrealistic motions (see Fig.~\ref{fig:intro}), the latent orthogonal space acts as a robust information bottleneck for learning meaningful motion representations, thereby aiding in achieving more accurate motion prediction. Thirdly, during the inference stage, controllable human motion prediction can be seamlessly achieved by manipulating the coefficients of the bases. Moreover, we made a surprising discovery that the learned bases already encode certain semantic information about human motion. For instance, as shown in Fig.~\ref{fig:intro}, the predicted motion can be guided towards different semantics by adjusting the coefficients. We therefore refer to this constructed latent space as \emph{\textbf{S}emantic \textbf{L}atent \textbf{D}irections}, or SLD in short. Expanding on SLD, we introduce a set of learnable motion queries. These queries facilitate the sampling of diverse motions, and when projected into the SLD space, more accurate, diverse, and coherent motion predictions are enabled.

In summary, our main contributions are threefold: 

1. We uncover that the latent motion space within current generative frameworks lacks the necessary constraints to effectively learn meaningful human motion representations for prediction.

2. We introduce a novel approach called Semantic Latent Directions (SLD), which constructs a latent semantic motion space, enabling accurate and controllable human motion prediction. Additionally, SLD is lightweight and can be seamlessly integrated into existing frameworks.

3. Through extensive experiments conducted on the HumanEva-I and Human3.6M datasets, we showcase that our method attains state-of-the-art performance in stochastic human motion prediction.

\section{Related Work}
\subsection{Human motion prediction}
Early works~\cite{DHMP1,DHMP2,DHMP3,DHMP4,DHMP5,DHMP6,DHMP7,Msrgcn,cues,gating,potr,Multilevel,Futurepong,multiscale} treat the HMP task as a deterministic regression problem that predicts a single human motion based on past human motion. While these methods fail in modeling the underlying multimodal distribution of the future motion, recent works~\cite{Dlow,MTVAE,DeLiGAN,DivSamp,GSPS,MOJO,STARS,HumanMac,MotionDiff} made efforts in predicting diverse future motions for each observed motion (\aka, stochastic human motion prediction or SHMP). To be specific, these methods assume that the hypothesis of underlying future motion obeys a latent distribution, and design generative models to learn such latent motion distribution, such as generative adversarial networks~\cite{DeLiGAN,GSPS}, variational autoencoders~\cite{Dlow,BOM,DivSamp,MOJO} and denoising diffusion probabilistic models~\cite{HumanMac,MotionDiff,Belfusion}. While the choice of generative models for SHMP is crucial, the specific properties of the latent motion distribution that are advantageous for SHMP remain unclear. Many existing works~\cite{Dlow,MTVAE,DeLiGAN,DivSamp,GSPS,MOJO,HumanMac,MotionDiff} incorporated a prior Gaussian distribution into the latent motion representation and focused on developing robust generative models or diverse sampling strategies. However, these approaches struggled to learn meaningful motion representations for accurate predictions in the absence of proper guidance on the latent distribution. Consequently, while they may achieve high diversity in motion predictions, they often generate inaccurate motions that are unrealistic and incoherent to past motion patterns (see Fig.~\ref{fig:intro}). A recent study, Belfusion~\cite{Belfusion}, has put forward an approach to disentangle human behavior within the generative latent space. Specifically, they extract behavior representations by combining future motion with the last three frames of past motion. Subsequently, a diffusion model is devised to capture the distribution of these behavior representations. This explicit representation imposes a strict constraint on the latent motion representation, leading to behavior-realistic predictions. Similarly, our goal in this paper is to attain a meaningful latent motion representation for precise SHMP. However, instead of enforcing explicit constraints like behavior representations, we define a set of semantic latent directions, enabling the model to autonomously learn these directions to imbue them with meaningful semantics.

\subsection{Controllable Motion Prediction}

Controllable human motion prediction has wide applications in the area of computer graphics, like virtual character control and games\cite{control1,control2,control3}. 
Previous works\cite{Dlow,GSPS,STARS,HumanMac,control_pr} mainly focus on low-level control on SHMP. For instance, DLow\cite{Dlow} and GSPS\cite{GSPS} control the generated future human motion by separating upper and lower body parts. HumanMAC\cite{HumanMac} achieves pose interpolation between two sequences. The recent Belfusion~\cite{Belfusion} involves human behaviors. While in this work, we propose to control SHMP in the latent semantic level for the first time. Benefit from the novel motion representation SLD, we could easily achieves this by editing the latent coefficients.

\subsection{Disentangled Representation Learning}
Our SLD is motivated by disentangled representation learning~\cite{higgins2016beta,burgess2018understanding,karras2019style,karras2020analyzing,karras2021alias,yang2023self,wang2021inmodegan}, an important task in generative modeling. For example, the well-known $\beta$-VAEs~\cite{higgins2016beta,burgess2018understanding} make a thorough study on training regularizations of VAEs to learn disentangled representations. 
StyleGANs~\cite{karras2019style,karras2020analyzing,karras2021alias} disentangle the style factors in the so-built latent $w$ space. Recent methods~\cite{yang2023disdiff,yang2024diffusion} also explore disentangled representation in diffusion models. In this paper, given the highly semantic and structured nature of human motion, we construct the latent disentangled space using Semantic Latent Directions (SLD), allowing us to achieve a disentangled semantic representation without the need for complex training regularizations.

\section{Methodology}

Given the past motion $X = [x_1, ..., x_{T_p}]$, where $T_p$ denote the sequence length, $x_i \in{R^{V \times 3}}$ represents the 3D human pose with $V$ joints. The objective of SHMP is to predict $K$ future motions $[\widehat{Y_1}, ..., \widehat{Y_K}]$ with length $T_f$. To ensure the accuracy and diversity of the predictions, one of the $\{\widehat{Y_i}\}_{i=1}^{K}$ should be as close as possible to the groundtruth $Y$ and the $K$ sequences should be diverse and realistic. 

In this section, we first review the general formulation of the human motion prediction task in \cref{sec:formulation}. Based on the general framework, we then in \cref{sec:SLD} introduce our core module \emph{Semantic Latent Directions} (SLD) equipped with diverse motion queries. At last, we summarize the training and inference process in \cref{sec:training}.

\subsection{General Formulation of SHMP} 
\label{sec:formulation}
The core problem of SHMP is to model the distribution $p(Y|X)$. It is non-trivial to directly parameterize this distribution with a neural network. To this end, existing methods usually introduce a latent variable $z$, assuming that the underlying future motion distribution can be derived from the latent distribution $p(z)$. The distribution $p(Y|X)$ can thus be re-parameterized as $p(Y|X) = \int p(Y|X, z)p(z)dz$, where $p(z)$ is usually assumed to be a Gaussian distribution and $p(Y|X, z)$ is implemented as a generator. The formulation of the above process can be summarized as follows:
\begin{align}
  z &\sim p(z), \label{eq:z} \\
  \widehat{Y} &= G_\phi(X, z), \label{eq:3-1}
\end{align}

The incorporation of the latent variable $z$ streamlines the motion prediction learning process.
While generative models theoretically have the potential to incorporate a diverse range of motion modes for $z$ to learn, there is no guarantee of accurately capturing all modes, as mode collapse is a frequently observed phenomenon. 
We posit that the Gaussian prior constraint on the latent distribution is insufficiently robust for acquiring meaningful hypothesis motion representations for 
prediction tasks. 
This limitation may lead models~\cite{Dlow,MOJO,DivSamp,STARS} to encounter difficulties in accurately capturing future motion patterns, and unrealistic predicted motions can often be observed in their approaches.

\begin{figure}[tb]
  \centering
  \includegraphics[width=\textwidth]{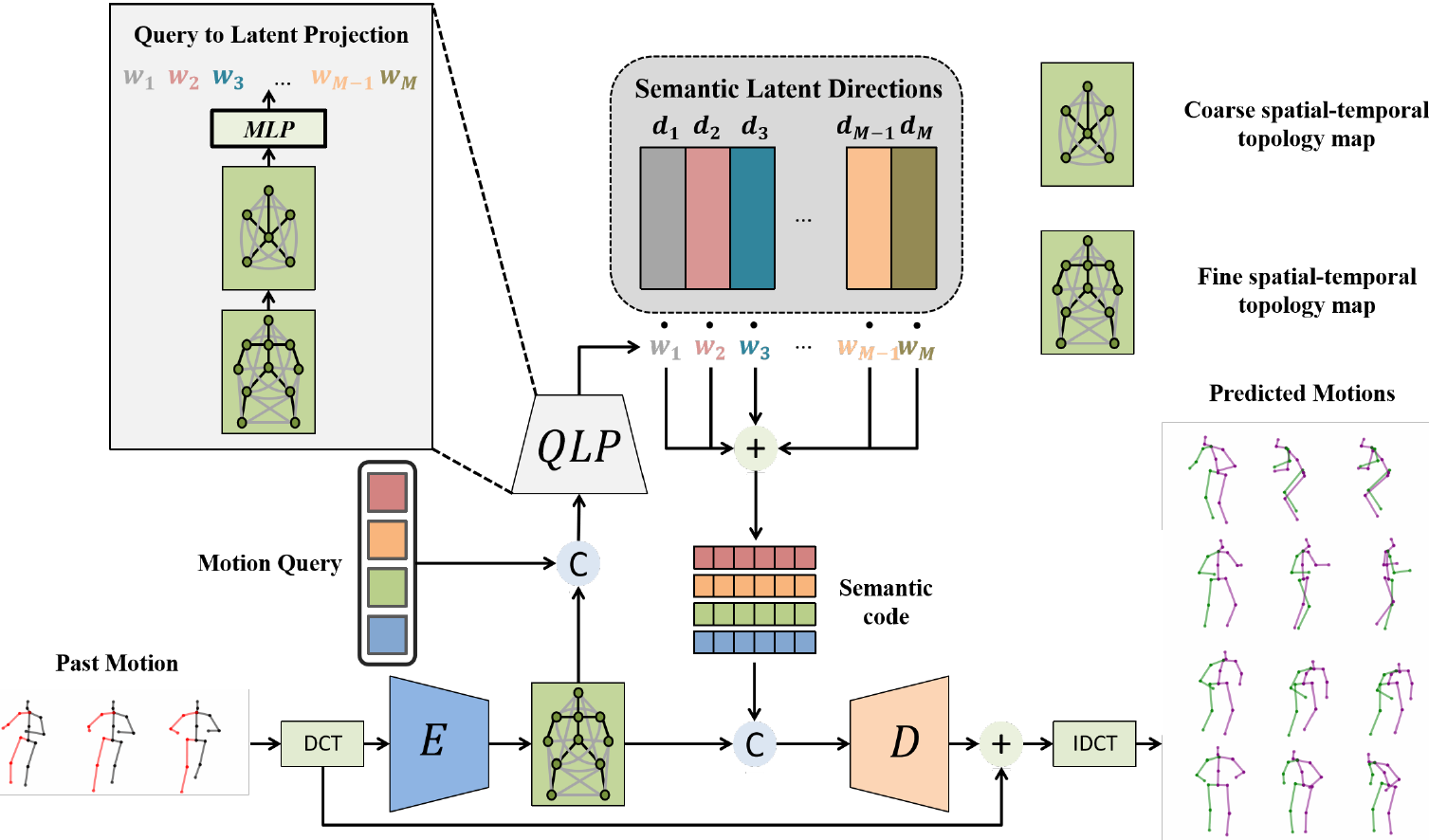}
  \caption{Overview of the framework: The past human motion is transformed to the frequency domain via DCT\cite{dct,GSPS}. The encoding feature of the past motion, along with the motion query, are merged and mapped into a series of latent coefficients $w = [w_1,..., w_M]$ through the Query to Latent Projection (QLP) module. Semantic codes are derived by integrating the semantic latent directions with the forecasted coefficients. Subsequently, the features of the past human motion and semantic codes are combined to predict future motion.}
  \label{fig:network}
\end{figure}

\subsection{Semantic Latent Space Modeling} 
\label{sec:SLD}

\subsubsection{Semantic Latent Directions.} 
To address above issues, instead of constraining the latent motion space with a Gaussian distribution, we build a semantic latent space with a set of semantic latent directions. 
Formally, denote $D = {[d_1,..., d_M]}\in R^{M\times C}$ as the $M$ latent directions which spans the latent motion space. 
We assume that the underlying latent factor of future motion $Y$ can be represented as a linear combination of these directions.
Denote the coefficients of the latent directions as $w = {[w_1,..., w_M]}$, we then formulate the 
process as follow:
\begin{eqnarray}
  z = \sum_{m=1}^{M}w_m\cdot d_m,\\
  \widehat{Y} = G_\phi(X, z), \label{eq:3-2}
\end{eqnarray}
In implementation, we set latent directions $D$ as learnable parameters and we predict $\{w_m\}_{m=1}^{M}$ from the past motion. To promote learning meaningful semantics for the latent directions, we further constrain $\{d_m\}_{m=1}^{M}$ to be orthogonal to each other, achieved by conducting $\operatorname{SVD}$ decomposition on the matrix $D$.

Intuitively, these latent directions operate similarly to semantic prototypes, where a human motion extracted from the dataset can be mapped onto these prototypes. For example, the motion "sit" could be depicted as a blend of the motions "stand" and "squat," as depicted in the upper segment of Fig.~\ref{fig:viz_l}. While mastering a Gaussian latent distribution to capture a spectrum of motion modes is challenging due to the common occurrence of model collapse, acquiring knowledge of the semantic latent directions and their corresponding coefficients proves to be significantly more manageable. This is attributed to two main factors. Firstly, we discretize the latent motion space into finite prototypes, mitigating abnormal predictions during training, as all forecasts must align with these prototypes. Secondly, by enforcing the orthogonality of the semantic latent directions, the learning complexity is further diminished.

In addition to enhancing the accurate modeling of the latent motion space, our method excels in enabling controllable human motion prediction through effortless adjustment of the coefficients linked to the latent directions. This stands in contrast to the challenges faced by traditional Gaussian-based latent spaces in achieving such control.

\begin{figure}[tb]
  \centering
  \includegraphics[width=\textwidth]{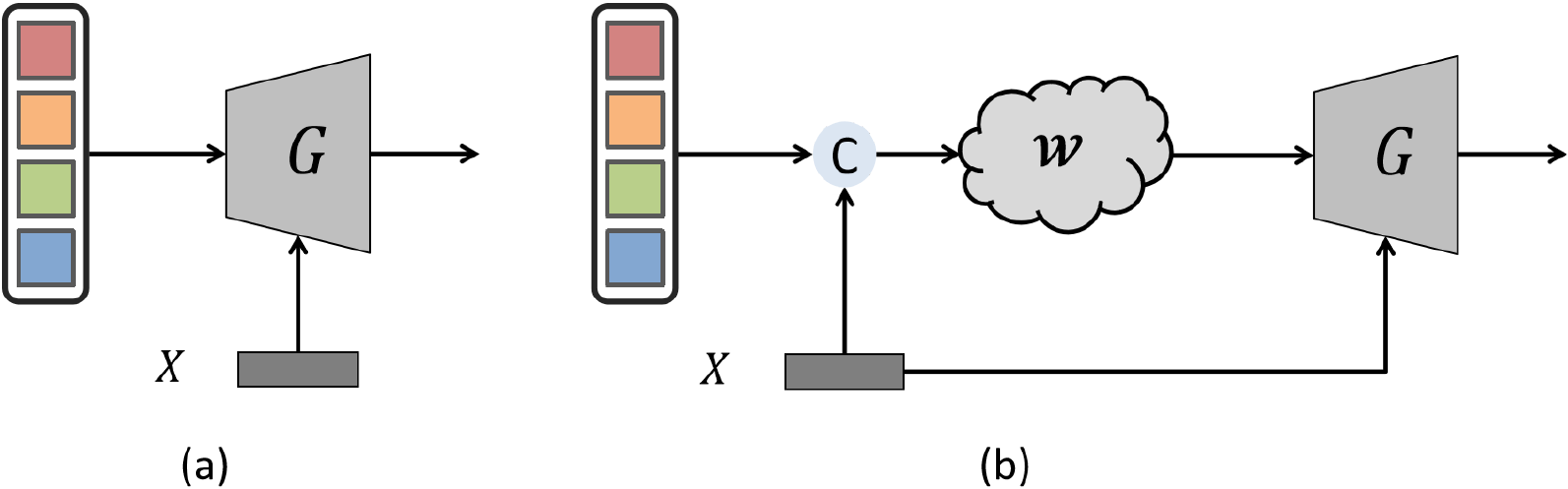}
  \caption{Illustration of different methods on promoting diverse motion sampling. (a) STARS~\cite{STARS} directly combined the motion queries with past motions. (b) We project the motion queries into latent coefficients of the semantic latent directions, ensuring accuracy during the diverse sampling.}
  \label{fig:query}
\end{figure}

\subsubsection{Diverse Motion Queries.} In order to encourage the generation of diverse motion predictions, current approaches~\cite{STARS,Dlow,DivSamp} either incorporate sub-modules to map the latent variable $z$ to multiple sub-distributions $z_1,...,z_K$, or directly incorporate $K$ trainable parameters to capture diverse motion modes. However, we contend that if the latent motion space is not appropriately constrained and learned, the subsequent diverse sampling strategies will lack significance. While these strategies may produce diverse motion samples, a considerable portion of them is likely to be unrealistic and inaccurate.

Fortunately, leveraging the modeling of our semantic latent space allows us to effortlessly achieve diverse predictions while maintaining accuracy. We thus introduce a set of adaptable \emph{motion queries} attached to the semantic latent directions. Instead of directly concatenating motion queries with past motion embeddings like~\cite{STARS}, we project these motion queries into the SLD space to derive varied coefficients for prediction. This projection into the SLD space ensures the precise prediction of each query. We depicted the various diverse sampling strategies in Fig.~\ref{fig:query}. Specifically, we denote the motion query set as $Q=[q_1,...,q_K]$, and the diverse motion prediction process can be formulated as follows:
\begin{eqnarray}
  w_{m}^k = \operatorname{Projection}(X, q_k),\\
  z_k = \sum_{m=1}^{M}w_m^k\cdot d_m,\\
  \widehat{Y_k} = G_\phi(X, z_k), k=1,...,K\label{eq:diverse}
\end{eqnarray}
The motion queries are randomly initialized and can be autonomously learned during training. We have observed that these motion queries can exhibit distinct motion patterns, as depicted in Fig.~\ref{fig:viz_a}. We will delve deeper into these observations in the later experimental sections.

\subsubsection{Overall Framework.} We implemented our Semantic Latent Directions (SLD) within a straightforward encoder-decoder framework. The entire architecture is depicted in Fig.~\ref{fig:network}. Specifically, past motion features are extracted through an encoder. To map motion queries to the coefficients of the semantic latent space, we introduce the query to latent projection network $\operatorname{QLP}$. The predicted semantic code is then combined with past motion features and fed into the decoder to generate the future pose sequence. Despite the simplicity of the framework, our SLD acts as a robust information bottleneck, allowing for a meaningful representation of the latent motion space and consequently enabling accurate, diverse, and controllable predictions. 

Notably, we follow previous work\cite{GSPS,STARS,HumanMac} in applying DCT/IDCT transform~\cite{dct,GSPS} to pre-process and post-process the past motion and predicted motion. 
To be specific, an input-agnostic projection matrix $F\in R^{(T_p+T_f) \times (T_p+T_f)}$ is pre-computed.
We project the past motion to the frequency domain with the first $N$-rows of $F$: $X_f=F[:N,:]X$, while back projecting the predicted frequency domain motion $\widehat{Y_f}$ to the time domain with the first $N$-columns of $F^{-1}$: $\widehat{Y}=F^{-1}[:,: N]\widehat{Y_f}$. Note that $X$ is padded to the length $T_p+T_f$ via repeating the last frame.

\subsection{Training and Inference}
\label{sec:training}
\subsubsection{Training.} Following existing approaches~\cite{GSPS,STARS} and for the sake of simplicity, we train the whole framework with three types of loss functions.
\begin{itemize}[itemsep=0pt]
    \item Reconstruction loss $\mathcal{L}_r$ between the GT future motion and the best prediction among the $K$ generated future motions, ensuring that the SLD can capture the accurate underlying future motion.
    \item Diversity-promoting loss $\mathcal{L}_d$ computes pairwise distances between $K$ generated future motions. This loss function encourages SLD to be complete to cover a wide range of motion modes and simultaneously to learn different patterns for the motion queries.
    \item Motion constraint loss $\mathcal{L}_c$ further constrains the predicted motion to be as reasonable as possible.
\end{itemize}
The overall objective function $\mathcal{L}$ can be summarized as follows:
\begin{equation}
  \mathcal{L} = \lambda_r\mathcal{L}_r + \lambda_d\mathcal{L}_d + \lambda_c\mathcal{L}_c,
  \label{eq:loss}
\end{equation}
where $\lambda_r$, $\lambda_{d}$ and the $\lambda_c$ are the weight of loss terms, we provide more details in the supplementary material. Note that we train the whole network end to end in one stage, the SLD is automatically learned together with the encoder and decoder.
\subsubsection{Inference.} During the inference stage, we input a past motion and produce $K$ basic motion predictions with latent coefficients $\{w_m^k\}_{m=1,...,M}^{k=1,...,K}$. To achieve controllable motion prediction, we edit the value of basic coefficients via simple adding value operations.

\section{Experiments}
\subsection{Experimental Setup for Stochastic HMP}
\subsubsection{Datasets.}
We evaluate the proposed SLD on two widely used motion capture datasets, Human3.6M\cite{h36m} and HumanEva-I\cite{humaneva}. The Human3.6M consists of 3.6 million frames recorded at 50 Hz, featuring 11 subjects performing 15 actions. Following previous works\cite{Dlow,GSPS,HumanMac,STARS}, the Human3.6M is split into the training set with subjects S1, S5, S6, S7, S8 and the testing set with subjects S9 and S11. We use 25 frames as the past motion to predict 100 future frames. The human motion of this dataset is represented as a sequence of 17-joint 3D poses similar to \cite{Dlow,GSPS,STARS}. The human motion of the HumanEva-I is represented as a sequence of 15-joint 3D poses\cite{Dlow,GSPS,STARS}. We adopt the official train/test split\cite{humaneva} and predict 60 future frames based on 15 past frames.

\subsubsection{Metrics.} Following previous works\cite{Dlow,DivSamp,HumanMac},
we employ the following five metrics to evaluate the diversity (APD\cite{apd}) and accuracy (ADE, FDE\cite{fde1,fde2}, MMADE and MMFDE) qualities of predicted motions. 
\begin{itemize}[itemsep=0pt]
    \item Average Pairwise Distance (APD) measures the diversity within samples by calculating the average $L_2$ distance between all pairs of motion samples, which is computed as $\frac{1}{K(K-1)}\sum_{i=1}^{K}\sum_{j\neq i}^K \left\|\ \widehat{Y_i} - \widehat{Y_j} \right\|\ $. 
    \item Average Displacement Error (ADE) computes the average $L_2$ distance between the GT future human motion $Y$ and the closest sample $\widehat{Y_i}$ over all time steps, which is computed as $\frac{1}{T_f}{min}_{\widehat{Y_i}\in \widehat{Y}}\left\|\ Y-\widehat{Y_i} \right\|\ $.
    \item Final Displacement Error (FDE) calculates the $L_2$ distance between the final GT pose and the closest sample's final pose, which is computed as ${min}_{\widehat{Y_i}\in \widehat{Y}}\left\|\ Y^{T_f}-{\widehat{Y_i}}^{T_f} \right\|\ $.
    \item Multi-Modal ADE (MMADE) and Multi-Modal FDE (MMFDE) are multi-modal versions of ADE and FDE, respectively, considering multi-modal GT\cite{MMADE} future motions, which are computed as $\frac{1}{MT_f}\sum_{m=1}^{M}min_{\widehat{Y_i}\in \widehat{Y}}\|\widehat{Y_i}-Y_m\|$ and $\frac{1}{M}\sum_{m=1}^N{min}_{\widehat{Y_i}\in \widehat{Y}}\left\|\ Y^{T_f}_m-{\widehat{Y_i}}^{T_f} \right\|\ $.
\end{itemize}

\subsubsection{Implementation Details.} The encoder and decoder consist of two STGCN layers\cite{stgcn1,stgcn2,stgcn3} and two pruned STGCN layers\cite{STARS}. The latent projection network $\operatorname{QLP}$ is implemented as three STGCN layers and three MLP layers. In each layer of GCN, we adopt batch normalization\cite{bs} and inject residual connections. We set $K=50$ following existing works~\cite{STARS,Dlow}. The model was trained for 500 epochs with a batch size of 16 on a single NVIDIA 3090 card with Pytorch\cite{pytorch}. We adopt the Adam\cite{adam} optimizer with the learning rate 0.001, which decayed according to training epochs as $lr=0.001 \times (1.0-\frac{max(0,epoch-100)}{400})$. It takes 25 hours for the training on the Human3.6M and 7 hours for that on the HumanEva-I. Additional implementation details are provided in the supplementary material. 

\subsubsection{Baselines.} To comprehensively evaluate our approach, we compare our method with a bunch of state-of-the-art baselines. Including ERD\cite{DHMP1}, acLSTM\cite{aclstm}, DeLiGAN\cite{DeLiGAN}, MT-CVAE\cite{MTVAE}, BoM\cite{BOM}, DivSamp\cite{DivSamp}, DLow\cite{Dlow}, MOJO\cite{MOJO}, STARS\cite{STARS}, Belfusion\cite{Belfusion}, MotionDiff\cite{MotionDiff} and HumanMac\cite{HumanMac}. Among them, ERD and acLSTM deterministic motion prediction methods, while the rest are stochastic motion prediction methods. In particular, approaches such as DLOW, HumanMAC, MotionDiff, and DivSamp have concentrated on constructing generative models directly within the human motion space to capture the underlying latent distribution. While Belfusion separates behavior from the latent motion representation.

We contend that without proper guidance on this latent distribution, deriving meaningful motion representations for subsequent predictive tasks becomes challenging, which may result in limited accuracy. The too-strict behavior representation (involving the concatenation of future frames with the last 3 frames of past motion) may still not offer a robust solution for prediction tasks. In contrast, our methodology defines a series of latent directions and empowers the model to organically imbue them with diverse semantics during training.

\begin{table}[tb]
\centering
\caption{Quantitative comparison between our approach and state-of-the-art methods on the HumanEva-I and Human3.6M datasets, our method consistently demonstrates superior accuracy while maintaining commendable diversity metrics.}
\label{tab:comparision}
\resizebox{\textwidth}{!}{%
\begin{tabular}{lccccclccccc}
\toprule
\multicolumn{1}{c}{\multirow{2}{*}{Method}}  & \multicolumn{5}{c}{HumanEva-I}          &  & \multicolumn{5}{c}{Human3.6M}        \\ \cmidrule(lr){2-6} \cmidrule(lr){7-12}
\multicolumn{1}{c}{} &
  APD$\uparrow$ &
  ADE$\downarrow$ &
  FDE$\downarrow$ &
  MMADE$\downarrow$ &
  MMFDE$\downarrow$ &
   &
  APD$\uparrow$ &
  ADE$\downarrow$ &
  FDE$\downarrow$ &
  MMADE$\downarrow$ &
  MMFDE$\downarrow$ \\ 
  \midrule
ERD\cite{DHMP1}       & 0 & 0.382 & 0.461 & 0.521 & 0.595 & & 0  & 0.722 & 0.969 & 0.776 & 0.995 \\
acLSTM\cite{aclstm}       & 0 & 0.429 & 0.541 & 0.530 & 0.608 & & 0  & 0.789 & 1.126 & 0.849 & 1.139 \\
DeLiGAN\cite{DeLiGAN}       & 2.177 & 0.306 & 0.322 & 0.385 & 0.371 & & 6.509  & 0.483 & 0.534 & 0.520 & 0.545 \\
BoM\cite{BOM}                & 2.846 & 0.271 & 0.279 & 0.373 & 0.351 & & 6.265  & 0.448 & 0.533 & 0.514 & 0.544\\
DLow\cite{Dlow}               & 4.855 & 0.251 & 0.268 & 0.362 & 0.339 & &  11.741 & 0.425 & 0.518 & 0.495 & 0.531\\
GSPS\cite{GSPS}               & 5.825 & 0.233 & 0.244 & 0.343 & 0.331 & & 14.757 & 0.389 & 0.496 & 0.476 & 0.525 \\
MOJO\cite{MOJO}               & 4.181 & 0.234 & 0.244 & 0.369 & 0.347 & & 12.579 & 0.412 & 0.514 & 0.497 & 0.538\\
DivSamp\cite{DivSamp}        & 6.109 & 0.220 & 0.234 & 0.342 & 0.316 & & 15.310 & 0.370 & 0.485 & 0.475 & 0.516\\
STARS\cite{STARS}            & 6.031 & 0.217 & 0.241 & 0.328 & 0.321 & & \bf 15.884 & 0.358 & 0.445 & 0.442 & 0.471\\
MotionDiff\cite{MotionDiff}  & 5.931 & 0.232 & 0.236 & 0.352 & 0.320 & &15.353 & 0.411 & 0.509 & 0.508 & 0.536\\
Belfusion\cite{Belfusion}    & -     & -     & -     & -     & -   &  & 7.602  & 0.372 & 0.474 & 0.473 & 0.507  \\
HumanMAC\cite{HumanMac}     & \bf 6.554 & 0.209 & 0.223 & 0.342 & 0.320 &  & 6.301  & 0.369 & 0.480 & 0.509 & 0.545\\ 
\midrule
SLD (Ours)                 & 4.066 & \bf 0.193 & \bf 0.209 & \bf 0.305 & \bf 0.293 &  & 8.741  & \bf 0.348 & \bf 0.436 & \bf 0.435 & \bf 0.463\\ 
\bottomrule
\end{tabular}%
}
\end{table}

\subsection{Quantitative Results}
\begin{figure}[tb]
  \centering
  \includegraphics[width=\textwidth]{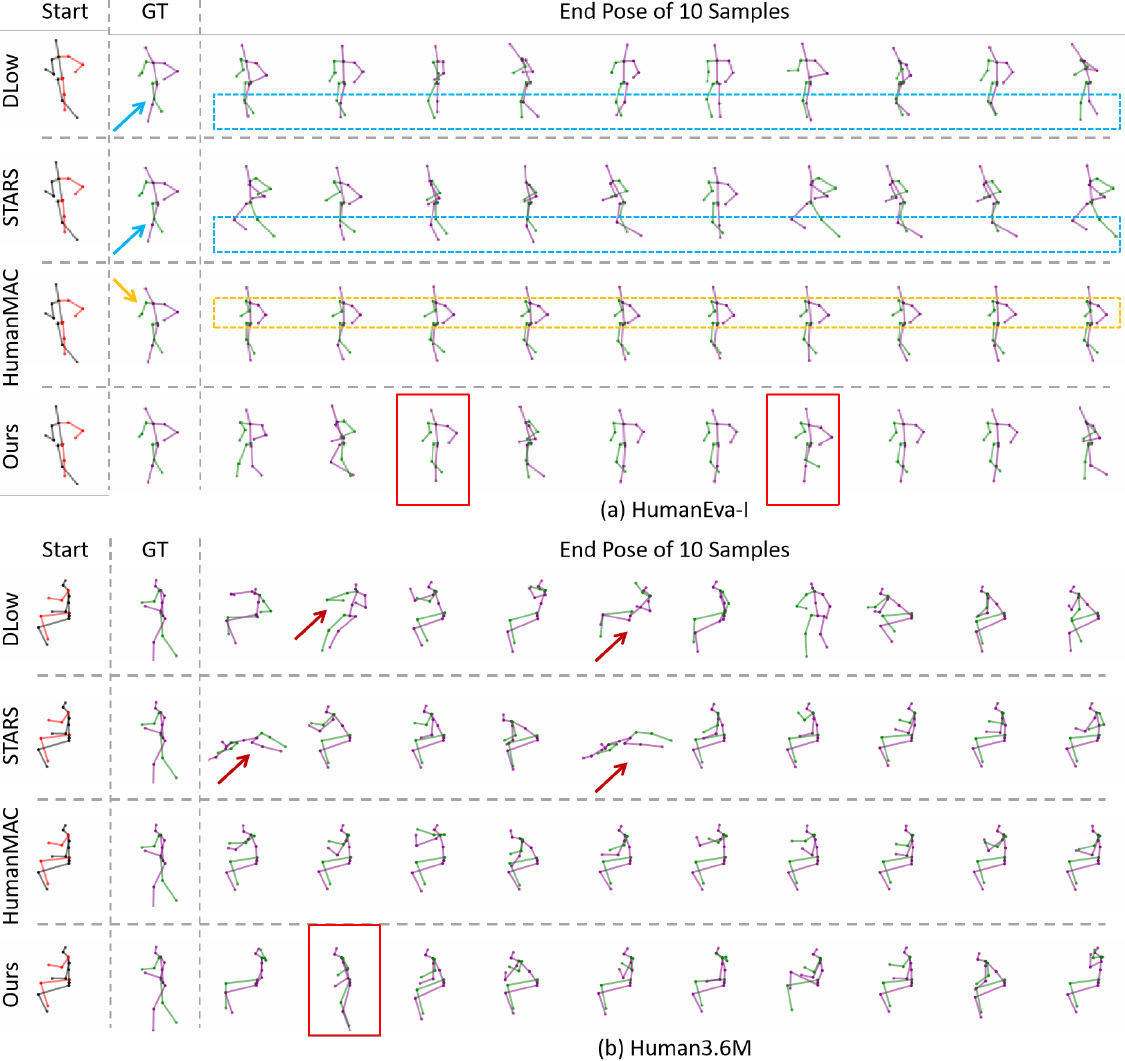}
  \caption{Qualitative comparison on Human3.6M and HumanEva-I datasets. We emphasize the accurate prediction with solid boxes while inaccurate and abnormal predictions are highlighted with dashed boxes and arrows. Our approach consistently demonstrates accurate, coherent, and diverse predictions.}
  \label{fig:viz_sota}
\end{figure}

We conducted a comparative analysis of our method against baseline approaches on the Human3.6M and HumanEva-I datasets, with the results outlined in \cref{tab:comparision}. Across all accuracy-based metrics, our approach significantly outperforms the previous state-of-the-art methods on both datasets. Additionally, we have achieved competitive performance in terms of diversity metrics. Notably, while methods like STARS~\cite{STARS} may exhibit higher diversity metrics such as APD, they often generate unrealistic motions, as illustrated in Fig.~\ref{fig:viz_sota}. In contrast, our method strikes a favorable balance between accurate and diverse predictions. 

Approaches that primarily focus on the generative modeling, such as DivSamp\cite{DivSamp}, HumanMAC\cite{HumanMac}, and MotionDiff\cite{MotionDiff}, exhibit notably lower accuracy levels (ADE, FDE, MMADE, and MMFDE), suggesting a lack of appropriate guidance in shaping the latent motion representation. Furthermore, despite Belfusion\cite{Belfusion} introducing a disentangled behavior representation, it still falls short of our method in terms of both accuracy and diversity. This reinforces the notion that our Semantic Latent Directions (SLD) serve as a more robust motion information bottleneck for human motion prediction.

\begin{table}[tb]
\centering
\caption{Ablation study on key components of our
approach. MQ denotes utilizing diverse motion queries only without SLD. MQ+SLD represents utilizing both SLD and diverse motion queries but without the query to SLD projection. MQ-P+SLD denotes our full model.}
\label{tab:ablation}
\resizebox{\textwidth}{!}{%
\begin{tabular}{lccccclccccc}
\toprule
\multicolumn{1}{c}{\multirow{2}{*}{}} &
  \multicolumn{5}{c}{HumanEva-I} &
   &
  \multicolumn{5}{c}{Human3.6M} \\ \cmidrule(lr){2-6} \cmidrule(lr){8-12} 
\multicolumn{1}{c}{} &
  APD$\uparrow$ &
  ADE$\downarrow$ &
  FDE$\downarrow$ &
  MMADE$\downarrow$ &
  MMFDE$\downarrow$ &
   &
  APD$\uparrow$ &
  ADE$\downarrow$ &
  FDE$\downarrow$ &
  MMADE$\downarrow$ &
  MMFDE$\downarrow$ \\ \midrule
MQ &
  1.562 &
  0.219 &
  0.248 &
  0.339 &
  0.339 &
   &
  7.286 &
  0.361 &
  0.449 &
  0.443 &
  0.475 \\
MQ+SLD  &
  3.365 &
  0.202 &
  0.218 &
  0.306 &
  0.294 &
   &
  7.936 &
  0.352 &
  0.442 &
  0.437 &
  0.468 \\
MQ-P+SLD &
   \bf 4.066 &
  \bf 0.193 &
  \bf 0.209 &
  \bf 0.305 &
  \bf 0.293 &
   &
  \bf 8.741 &
  \bf 0.348 &
  \bf 0.436 &
  \bf 0.435 &
  \bf 0.463 \\ 
 \bottomrule
\end{tabular}%
}
\end{table}

\subsection{Component Ablations}
We here meticulously assess the impact of the fundamental components of our model, specifically focusing on the proposed Semantic Latent Directions (SLD) and the diverse motion queries associated with them. The results are summarized in \cref{tab:ablation}. Initially, by utilizing motion queries in isolation without the SLD, a notable decrease in performance is observed compared to our full model. Upon integrating the SLD, a considerable enhancement is observed. Furthermore, projecting the motion queries into the SLD space leads to even greater performance improvements, ultimately yielding the best results. These outcomes underscore the efficacy of our SLD framework and emphasize how the SLD, in conjunction with diverse motion queries, can significantly elevate the overall performance.

\begin{figure}[tb]
  \centering
  \includegraphics[width=\textwidth]{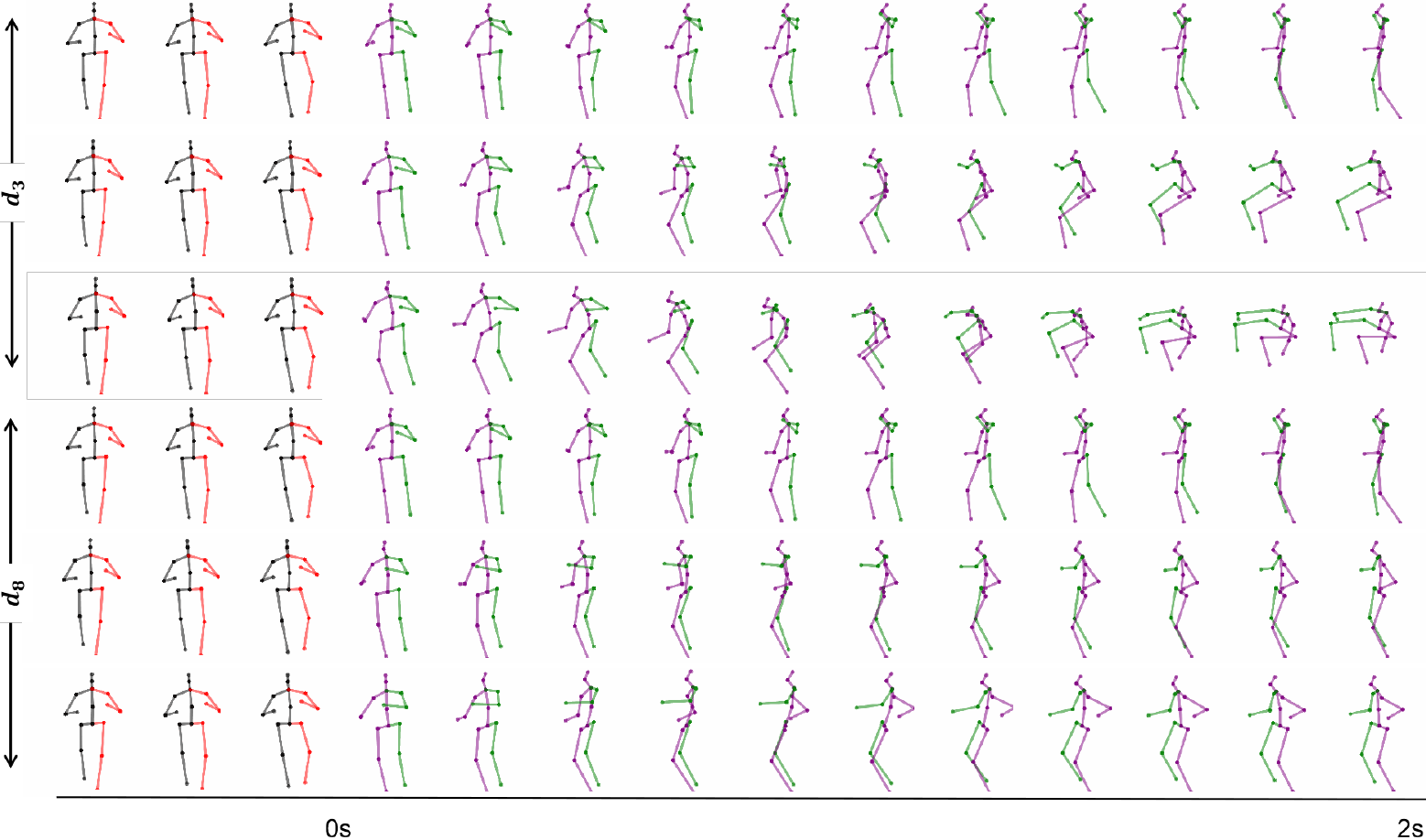}
  \caption{Visualization of controllable motion prediction on the Human3.6M. Semantic control can be achieved by adjusting the coefficients in specific directions. Different degrees of semantic alterations can be attained by varying the magnitude of the coefficient change.}
  \label{fig:viz_l}
\end{figure}

\begin{figure}[tb]
  \centering
  \includegraphics[width=\textwidth]{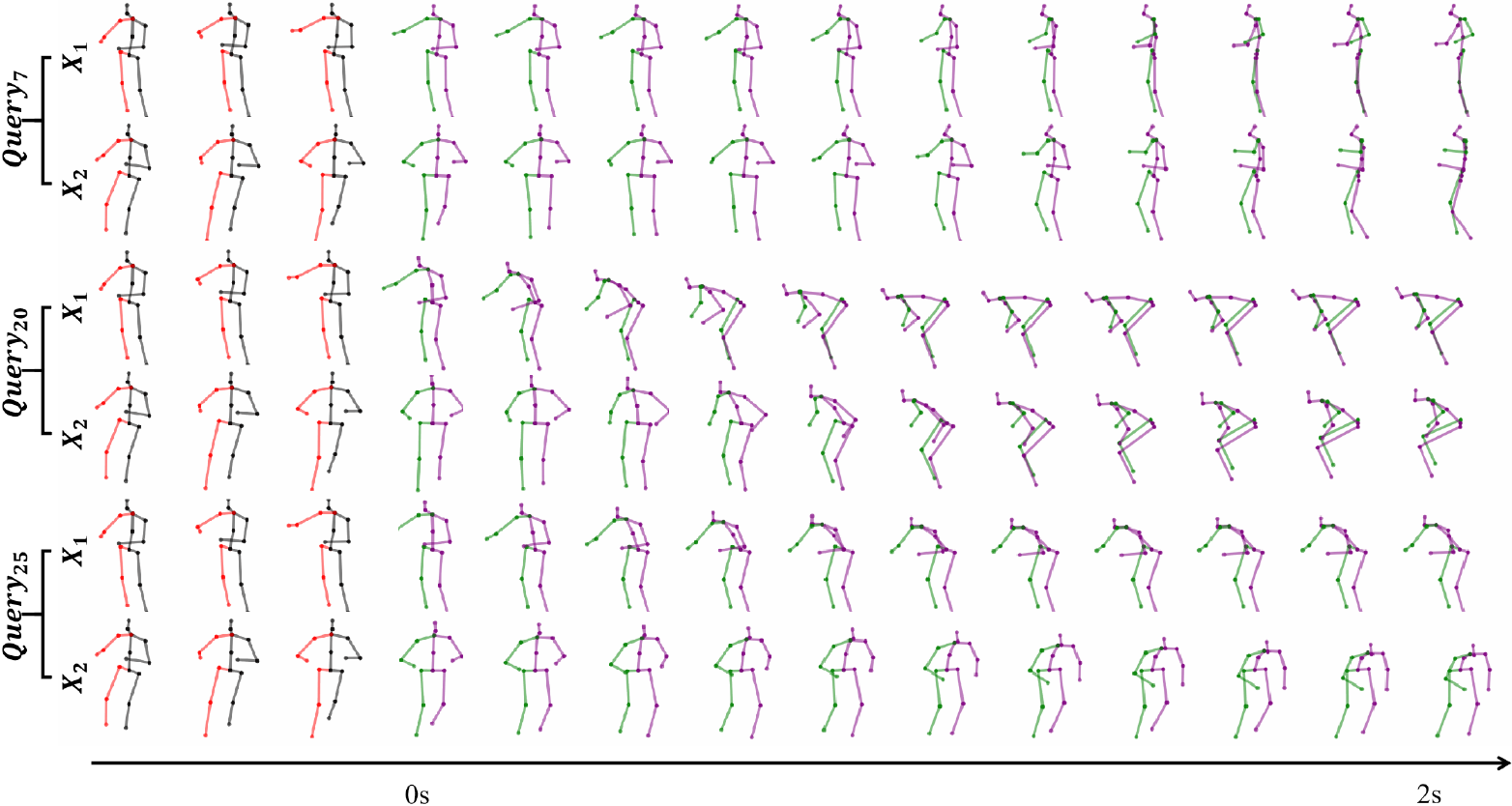}
  \caption{Visualization of motion patterns captured by motion queries on the Human3.6M dataset, showcases the ability of different motion queries to accurately capture a diverse range of motion patterns.}
  \label{fig:viz_a}
\end{figure}

\subsection{Qualitative Results}

\noindent\textbf{Qualitative Comparison.} We further showcase qualitative comparisons in Fig.~\ref{fig:viz_sota}. The visualization includes the initial poses of past human motions, the GT end poses, and the predicted end poses generated by various methods across 10 samples. Despite our meticulous selection of optimal outcomes from the baseline methods, they still exhibit subpar accuracy, with fewer predicted poses aligning closely with the GT and occasional abnormal poses, particularly noticeable in the results from STARS.

In contrast, our approach consistently demonstrates the ability to accurately capture end poses while preserving a decent level of diversity. Additionally, our projected motions exhibit a greater degree of naturalness and coherence with past human movements, showcasing minimal deviation from historical motion patterns. These findings validate the efficacy of our SLD in effectively capturing future motions while upholding diversity in predictions.

\noindent\textbf{Controllable Motion Prediction.} We can effortlessly facilitate controllable motion prediction by manipulating the coefficients associated with the SLD. In this context, we showcase the outcomes of editing two specific directions in Fig.~\ref{fig:viz_l}. Notably, these results illustrate how adjusting the coefficients enables the control of transitions from standing to sitting and squatting, as well as the modulation of arm swing amplitudes and directions.

These observations confirm that our SLD framework enables a novel form of semantic controllable motion prediction, offering a versatile and intuitive approach to directing predicted motions.

\noindent\textbf{Diverse Motion Sampling.} The motion queries linked with our semantic latent directions have been adeptly trained to encapsulate diverse motion patterns. The versatility is evident in Fig.~\ref{fig:viz_a}, where distinct motion queries exhibit a propensity for capturing various motion patterns such as turning left and right, sitting, squatting, and more. This capability underscores the adaptability of our SLD associated with motion queries in encompassing human movements with high precision and fidelity.

\section{Conclusion}

In this paper, we introduce a novel method called Semantic Latent Directions (SLD) for stochastic human motion prediction. SLD defines a series of orthogonal latent directions, excelling in capturing meaningful motion semantics and enhancing the accuracy of motion predictions. Additionally, SLD offers controllable prediction capabilities by manipulating its coefficients during the inference phase. Expanding on SLD, we further introduce a spectrum of diverse motion queries. By aligning these motion queries with the SLD space, our approach is enriched, resulting in coherent and diverse motion sampling outcomes. Extensive experiments on widely used benchmarks validate the superiority of our method in accurately predicting motions while maintaining decent realism and diversity. We believe that our approach has great potential in learning disentangled motion representations.

\section{Limitation and future work.} 
In this paper, we focus on constraining the latent motion space using latent semantic directions within a controlled setting for stochastic human motion prediction. However, exploring scenarios where humans interact with their environments to determine their movements presents an intriguing avenue for future research. Investigating semantic motion representation in such dynamic and interactive contexts could offer valuable insights and pose an interesting problem worth studying.\\

\section*{Acknowledgement.} This work was partially supported by the National Natural Science Foundation of China (No. 62176047), the Shenzhen Fundamental Research Program (No. JCYJ20220530164812027), the Shenzhen Science and Technology Plan Project (No. KJZD20230923113800002) and the Shenzhen LongHua Fundamental Research Program (No. 10162A20230325B73A546).

\clearpage  

%
%
\bibliographystyle{splncs04}
\bibliography{main}

\begin{thebibliography}{10}
\providecommand{\url}[1]{\texttt{#1}}
\providecommand{\urlprefix}{URL }
\providecommand{\doi}[1]{https://doi.org/#1}

\bibitem{DHMP5}
Aksan, E., Kaufmann, M., Cao, P., Hilliges, O.: A spatio-temporal transformer for 3d human motion prediction. In: 2021 International Conference on 3D Vision (3DV). pp. 565--574. IEEE (2021)

\bibitem{apd}
Aliakbarian, S., Saleh, F.S., Salzmann, M., Petersson, L., Gould, S.: A stochastic conditioning scheme for diverse human motion prediction. In: Proceedings of the IEEE/CVF Conference on Computer Vision and Pattern Recognition. pp. 5223--5232 (2020)

\bibitem{Belfusion}
Barquero, G., Escalera, S., Palmero, C.: Belfusion: Latent diffusion for behavior-driven human motion prediction. In: Proceedings of the IEEE/CVF International Conference on Computer Vision. pp. 2317--2327 (2023)

\bibitem{BOM}
Bhattacharyya, A., Schiele, B., Fritz, M.: Accurate and diverse sampling of sequences based on a “best of many” sample objective. In: Proceedings of the IEEE Conference on Computer Vision and Pattern Recognition. pp. 8485--8493 (2018)

\bibitem{burgess2018understanding}
Burgess, C.P., Higgins, I., Pal, A., Matthey, L., Watters, N., Desjardins, G., Lerchner, A.: Understanding disentangling in $beta$-vae. arXiv preprint arXiv:1804.03599  (2018)

\bibitem{interaction1}
B{\"u}tepage, J., Kjellstr{\"o}m, H., Kragic, D.: Anticipating many futures: Online human motion prediction and generation for human-robot interaction. In: 2018 IEEE international conference on robotics and automation (ICRA). pp. 4563--4570. IEEE (2018)

\bibitem{DHMP7}
Cai, Y., Huang, L., Wang, Y., Cham, T.J., Cai, J., Yuan, J., Liu, J., Yang, X., Zhu, Y., Shen, X., et~al.: Learning progressive joint propagation for human motion prediction. In: Computer Vision--ECCV 2020: 16th European Conference, Glasgow, UK, August 23--28, 2020, Proceedings, Part VII 16. pp. 226--242. Springer (2020)

\bibitem{HumanMac}
Chen, L.H., Zhang, J., Li, Y., Pang, Y., Xia, X., Liu, T.: Humanmac: Masked motion completion for human motion prediction. arXiv preprint arXiv:2302.03665  (2023)

\bibitem{Msrgcn}
Dang, L., Nie, Y., Long, C., Zhang, Q., Li, G.: Msr-gcn: Multi-scale residual graph convolution networks for human motion prediction. In: Proceedings of the IEEE/CVF International Conference on Computer Vision. pp. 11467--11476 (2021)

\bibitem{DivSamp}
Dang, L., Nie, Y., Long, C., Zhang, Q., Li, G.: Diverse human motion prediction via gumbel-softmax sampling from an auxiliary space. In: Proceedings of the 30th ACM International Conference on Multimedia. pp. 5162--5171 (2022)

\bibitem{DHMP1}
Fragkiadaki, K., Levine, S., Felsen, P., Malik, J.: Recurrent network models for human dynamics. In: Proceedings of the IEEE international conference on computer vision. pp. 4346--4354 (2015)

\bibitem{DHMP6}
Gao, X., Du, S., Wu, Y., Yang, Y.: Decompose more and aggregate better: Two closer looks at frequency representation learning for human motion prediction. In: Proceedings of the IEEE/CVF Conference on Computer Vision and Pattern Recognition. pp. 6451--6460 (2023)

\bibitem{control_pr}
Gu, C., Yu, J., Zhang, C.: Learning disentangled representations for controllable human motion prediction. Pattern Recognition  \textbf{146},  109998 (2024)

\bibitem{DHMP2}
Gui, L.Y., Wang, Y.X., Liang, X., Moura, J.M.: Adversarial geometry-aware human motion prediction. In: Proceedings of the european conference on computer vision (ECCV). pp. 786--803 (2018)

\bibitem{fde1}
Gupta, A., Johnson, J., Fei-Fei, L., Savarese, S., Alahi, A.: Social gan: Socially acceptable trajectories with generative adversarial networks. In: Proceedings of the IEEE conference on computer vision and pattern recognition. pp. 2255--2264 (2018)

\bibitem{DeLiGAN}
Gurumurthy, S., Kiran~Sarvadevabhatla, R., Venkatesh~Babu, R.: Deligan: Generative adversarial networks for diverse and limited data. In: Proceedings of the IEEE conference on computer vision and pattern recognition. pp. 166--174 (2017)

\bibitem{higgins2016beta}
Higgins, I., Matthey, L., Pal, A., Burgess, C., Glorot, X., Botvinick, M., Mohamed, S., Lerchner, A.: beta-vae: Learning basic visual concepts with a constrained variational framework. In: International conference on learning representations (2016)

\bibitem{control1}
Holden, D., Komura, T., Saito, J.: Phase-functioned neural networks for character control. ACM Transactions on Graphics (TOG)  \textbf{36}(4),  1--13 (2017)

\bibitem{control2}
Holden, D., Saito, J., Komura, T.: A deep learning framework for character motion synthesis and editing. ACM Transactions on Graphics (TOG)  \textbf{35}(4),  1--11 (2016)

\bibitem{bs}
Ioffe, S., Szegedy, C.: Batch normalization: Accelerating deep network training by reducing internal covariate shift. In: International conference on machine learning. pp. 448--456. pmlr (2015)

\bibitem{h36m}
Ionescu, C., Papava, D., Olaru, V., Sminchisescu, C.: Human3. 6m: Large scale datasets and predictive methods for 3d human sensing in natural environments. IEEE transactions on pattern analysis and machine intelligence  \textbf{36}(7),  1325--1339 (2013)

\bibitem{DHMP3}
Jain, A., Zamir, A.R., Savarese, S., Saxena, A.: Structural-rnn: Deep learning on spatio-temporal graphs. In: Proceedings of the ieee conference on computer vision and pattern recognition. pp. 5308--5317 (2016)

\bibitem{karras2021alias}
Karras, T., Aittala, M., Laine, S., H{\"a}rk{\"o}nen, E., Hellsten, J., Lehtinen, J., Aila, T.: Alias-free generative adversarial networks. Advances in Neural Information Processing Systems  \textbf{34},  852--863 (2021)

\bibitem{karras2019style}
Karras, T., Laine, S., Aila, T.: A style-based generator architecture for generative adversarial networks. In: Proceedings of the IEEE/CVF conference on computer vision and pattern recognition. pp. 4401--4410 (2019)

\bibitem{karras2020analyzing}
Karras, T., Laine, S., Aittala, M., Hellsten, J., Lehtinen, J., Aila, T.: Analyzing and improving the image quality of stylegan. In: Proceedings of the IEEE/CVF conference on computer vision and pattern recognition. pp. 8110--8119 (2020)

\bibitem{adam}
Kingma, D.P., Ba, J.: Adam: A method for stochastic optimization. arXiv preprint arXiv:1412.6980  (2014)

\bibitem{koppula2015anticipating}
Koppula, H.S., Saxena, A.: Anticipating human activities using object affordances for reactive robotic response. IEEE transactions on pattern analysis and machine intelligence  \textbf{38}(1),  14--29 (2015)

\bibitem{interaction2}
Koppula, H.S., Saxena, A.: Anticipating human activities using object affordances for reactive robotic response. IEEE transactions on pattern analysis and machine intelligence  \textbf{38}(1),  14--29 (2015)

\bibitem{interaction3}
Lasota, P.A., Shah, J.A.: A multiple-predictor approach to human motion prediction. In: 2017 IEEE International Conference on Robotics and Automation (ICRA). pp. 2300--2307. IEEE (2017)

\bibitem{fde2}
Lee, N., Choi, W., Vernaza, P., Choy, C.B., Torr, P.H., Chandraker, M.: Desire: Distant future prediction in dynamic scenes with interacting agents. In: Proceedings of the IEEE conference on computer vision and pattern recognition. pp. 336--345 (2017)

\bibitem{aclstm}
Li, X., Li, H., Joo, H., Liu, Y., Sheikh, Y.: Structure from recurrent motion: From rigidity to recurrency. In: Proceedings of the IEEE conference on computer vision and pattern recognition. pp. 3032--3040 (2018)

\bibitem{control3}
Ling, H.Y., Zinno, F., Cheng, G., Van De~Panne, M.: Character controllers using motion vaes. ACM Transactions on Graphics (TOG)  \textbf{39}(4),  40--1 (2020)

\bibitem{cues}
Liu, Z., Su, P., Wu, S., Shen, X., Chen, H., Hao, Y., Wang, M.: Motion prediction using trajectory cues. In: Proceedings of the IEEE/CVF international conference on computer vision. pp. 13299--13308 (2021)

\bibitem{luber2010people}
Luber, M., Stork, J.A., Tipaldi, G.D., Arras, K.O.: People tracking with human motion predictions from social forces. In: 2010 IEEE international conference on robotics and automation. pp. 464--469. IEEE (2010)

\bibitem{GSPS}
Mao, W., Liu, M., Salzmann, M.: Generating smooth pose sequences for diverse human motion prediction. In: Proceedings of the IEEE/CVF International Conference on Computer Vision. pp. 13309--13318 (2021)

\bibitem{dct}
Mao, W., Liu, M., Salzmann, M., Li, H.: Learning trajectory dependencies for human motion prediction. In: Proceedings of the IEEE/CVF international conference on computer vision. pp. 9489--9497 (2019)

\bibitem{Multilevel}
Mao, W., Liu, M., Salzmann, M., Li, H.: Multi-level motion attention for human motion prediction. International journal of computer vision  \textbf{129}(9),  2513--2535 (2021)

\bibitem{DHMP4}
Martinez, J., Black, M.J., Romero, J.: On human motion prediction using recurrent neural networks. In: Proceedings of the IEEE conference on computer vision and pattern recognition. pp. 2891--2900 (2017)

\bibitem{potr}
Mart{\'\i}nez-Gonz{\'a}lez, A., Villamizar, M., Odobez, J.M.: Pose transformers (potr): Human motion prediction with non-autoregressive transformers. In: Proceedings of the IEEE/CVF International Conference on Computer Vision. pp. 2276--2284 (2021)

\bibitem{paden2016survey}
Paden, B., {\v{C}}{\'a}p, M., Yong, S.Z., Yershov, D., Frazzoli, E.: A survey of motion planning and control techniques for self-driving urban vehicles. IEEE Transactions on intelligent vehicles  \textbf{1}(1),  33--55 (2016)

\bibitem{pytorch}
Paszke, A., Gross, S., Massa, F., Lerer, A., Bradbury, J., Chanan, G., Killeen, T., Lin, Z., Gimelshein, N., Antiga, L., et~al.: Pytorch: An imperative style, high-performance deep learning library. Advances in neural information processing systems  \textbf{32} (2019)

\bibitem{humaneva}
Sigal, L., Balan, A.O., Black, M.J.: Humaneva: Synchronized video and motion capture dataset and baseline algorithm for evaluation of articulated human motion. International journal of computer vision  \textbf{87}(1-2),  4--27 (2010)

\bibitem{stgcn1}
Sofianos, T., Sampieri, A., Franco, L., Galasso, F.: Space-time-separable graph convolutional network for pose forecasting. In: Proceedings of the IEEE/CVF International Conference on Computer Vision. pp. 11209--11218 (2021)

\bibitem{character_animation}
Starke, S., Zhao, Y., Zinno, F., Komura, T.: Neural animation layering for synthesizing martial arts movements. ACM Transactions on Graphics (TOG)  \textbf{40}(4),  1--16 (2021)

\bibitem{healthcare}
Troje, N.F.: Decomposing biological motion: A framework for analysis and synthesis of human gait patterns. Journal of vision  \textbf{2}(5), ~2--2 (2002)

\bibitem{wang2021inmodegan}
Wang, Y., Bremond, F., Dantcheva, A.: Inmodegan: Interpretable motion decomposition generative adversarial network for video generation. arXiv preprint arXiv:2101.03049  (2021)

\bibitem{MotionDiff}
Wei, D., Sun, H., Li, B., Lu, J., Li, W., Sun, X., Hu, S.: Human joint kinematics diffusion-refinement for stochastic motion prediction. In: Proceedings of the AAAI Conference on Artificial Intelligence. vol.~37, pp. 6110--6118 (2023)

\bibitem{interaction4}
Wu, E., Koike, H.: Futurepong: Real-time table tennis trajectory forecasting using pose prediction network. In: Extended Abstracts of the 2020 CHI Conference on Human Factors in Computing Systems. pp.~1--8 (2020)

\bibitem{Futurepong}
Wu, E., Koike, H.: Futurepong: Real-time table tennis trajectory forecasting using pose prediction network. In: Extended Abstracts of the 2020 CHI Conference on Human Factors in Computing Systems. pp.~1--8 (2020)

\bibitem{STARS}
Xu, S., Wang, Y.X., Gui, L.Y.: Diverse human motion prediction guided by multi-level spatial-temporal anchors. In: European Conference on Computer Vision. pp. 251--269. Springer (2022)

\bibitem{MTVAE}
Yan, X., Rastogi, A., Villegas, R., Sunkavalli, K., Shechtman, E., Hadap, S., Yumer, E., Lee, H.: Mt-vae: Learning motion transformations to generate multimodal human dynamics. In: Proceedings of the European conference on computer vision (ECCV). pp. 265--281 (2018)

\bibitem{stgcn2}
Yan, Z., Zhai, D.H., Xia, Y.: Dms-gcn: dynamic mutiscale spatiotemporal graph convolutional networks for human motion prediction. arXiv preprint arXiv:2112.10365  (2021)

\bibitem{yang2023self}
Yang, D., Wang, Y., Kong, Q., Dantcheva, A., Garattoni, L., Francesca, G., Bremond, F.: Self-supervised video representation learning via latent time navigation. arXiv preprint arXiv:2305.06437  (2023)

\bibitem{yang2024diffusion}
Yang, T., Lan, C., Lu, Y., et~al.: Diffusion model with cross attention as an inductive bias for disentanglement. arXiv preprint arXiv:2402.09712  (2024)

\bibitem{yang2023disdiff}
Yang, T., Wang, Y., Lu, Y., Zheng, N.: Disdiff: Unsupervised disentanglement of diffusion probabilistic models. In: Thirty-seventh Conference on Neural Information Processing Systems (2023), \url{https://openreview.net/forum?id=3ofe0lpwQP}

\bibitem{stgcn3}
Yu, B., Yin, H., Zhu, Z.: Spatio-temporal graph convolutional networks: A deep learning framework for traffic forecasting. arXiv preprint arXiv:1709.04875  (2017)

\bibitem{MMADE}
Yuan, Y., Kitani, K.: Diverse trajectory forecasting with determinantal point processes. arXiv preprint arXiv:1907.04967  (2019)

\bibitem{Dlow}
Yuan, Y., Kitani, K.: Dlow: Diversifying latent flows for diverse human motion prediction. In: Computer Vision--ECCV 2020: 16th European Conference, Glasgow, UK, August 23--28, 2020, Proceedings, Part IX 16. pp. 346--364. Springer (2020)

\bibitem{multiscale}
Zand, M., Etemad, A., Greenspan, M.: Multiscale residual learning of graph convolutional sequence chunks for human motion prediction. arXiv preprint arXiv:2308.16801  (2023)

\bibitem{MOJO}
Zhang, Y., Black, M.J., Tang, S.: We are more than our joints: Predicting how 3d bodies move. In: Proceedings of the IEEE/CVF Conference on Computer Vision and Pattern Recognition. pp. 3372--3382 (2021)

\bibitem{gating}
Zhong, C., Hu, L., Zhang, Z., Ye, Y., Xia, S.: Spatio-temporal gating-adjacency gcn for human motion prediction. In: Proceedings of the IEEE/CVF Conference on Computer Vision and Pattern Recognition. pp. 6447--6456 (2022)

\end{thebibliography}
\end{document}